# Image Hash Minimization for Tamper Detection

Subhajit Maity
*Department of ECE*
*Jalpaiguri Govt. Engg. College*
Jalpaiguri, India
smaity.jgec18@gmail.com

Ram Kumar Karsh
*Department of ECE*
*NIT Silchar*
Assam, India
tnramkarsh@gmail.com

*Abstract*—Tamper detection using image hash is a very common problem of modern days. Several research and advancements have already been done to address this problem. However, most of the existing methods lack the accuracy of tamper detection when the tampered area is low, as well as requiring long image hashes. In this paper, we propose a novel method objectively to minimize the hash length while enhancing the performance at low tampered area.

## I. INTRODUCTION

Today is the age of large data handling. According to Flickr statistics, more than one million photos are shared everyday on the Flickr website. No one has enough time to keep track of all these data. On the other hand, with growing online marketing and e-commerce, information security is becoming more and more important. Image is one of the most popular modes of sharing data and transmitting information in modern days. With the emergence of several digital image manipulation technologies, the authenticity of an image during transmission is compromised in a great scale.

In the past decade, several researchers had tried to earn the trust back and to provide several methods to check the authenticity of a digital image. Generally, there are two types of methods by which tampering is done. One is copy move forgery, which can be detected with in a tampered image without any resource taken from the original image. Another is any arbitrary tampering of the image. Now for detection of this type of tampering two types of methods are used in modern days, namely Watermarking and Image Hashing.

Watermarking is a very popular method used for detecting if an image is authentic or not. Digital watermarking is a very popular application of image processing in which information is hidden within an image using a key which covers the hidden information in the image. The key is used to extract the information later on which can be used for detecting if the image is tampered or not. There are several methods for watermarking like the one presented in [1]. Watermarking is quite popular though it has a drawback. The procedure of embedding the watermark visually degrades the image in many folds. Hence for the applications which demand high visual quality watermarking is not a desired solution.

To solve this problem, image hashes is the newest research approach on image tampering detection. Venkatesan *et al.* [2] first introduced the idea of image hashing. Image hashes are the compact signatures or 'fingerprints' for an image which should be robust enough to survive several content preserving operations as well as should be effective enough to detect any malicious image manipulations. There are several other available methods for tamper detection, among which the one using core alignment presented by Ma *et al.* [3] is quite popular. Kozat *et al.* [4] also has suggested a method to extract image hashes by the method of Singular Value Decomposition (SVD). A new dimension reduction method named Non-negative Matrix Factorization (NMF) [5], [6] has also been developed for tamper detection, which further improved the existing method against a wide range of insignificant attacks using a small hashing length; however, the method suffered from brightness change as well as large geometrical change. Then again Roy and Sun [7] developed the method of detecting the area where the tampering is done. There are several other existing methods like the wavelet-based hashing method, suggested by Ahmed *et al.* [8] as well as the Quaternion based image hashing method proposed by Yan *et al.* [9], which is quite robust indeed.

Monga *et al.* [10] has shown that salient feature points can be used to generate image hash, although even if the method was able to survive several standard benchmark attacks, the matching accuracy is quite low. Ouyang *et al.* [11] and Zhao *et al.* [12] has also proposed methods for tamper detection and copy move forgery localization which use Zernike moments as well as local features. Shaikh and Sonavane [13] proposed a new approach which just segments the image into several rings robust to geometric transformations and then detects the features along the ring to create the image hash. Battiato *et al.* [14] has given a new approach of using spatial distribution of the features as image hash. Lu and Wu [15] used SIFT features to generate the image hash. However, Lv and Wang [16] suggested the shape context-based image hashing method using the SIFT-Harris detector. Tang *et al.* [17] has suggested the method of image hashing using the color vector angle and image edges found by Canny edge operator. Wang *et al.* [18] has developed a method for image hashing working on block based and key-point based features which were achieving the goal quite well, but the hash is too long. Several other papers [20], [21], [22], [23], [24] have already shown that image hashes can be generated for tamper detection.

The most of the existing methods can detect tampering with quite a low accuracy with a compact hash. Even when the various content preserving operations are taken into account, the accuracy decreases very steeply. Moreover, to detect tampering where tampered area is as low as 5% of the image size, no attempt has been done till date as far of our knowledge. To resolve this problem, we have presented a completely novel approach for tamper detection.

Firstly, we are using Speeded Up Robust Features (SURF) for extracting the image features which are robust against scale change and distortion or deformations. The image hash is not directly generated using the features.

Secondly, along with the feature extraction, we kept in mind that the hash length should be as low as possible, and hence hash length minimization is also done in this paper. The hash is generated by applying K-Means clustering on the SURF features acquired from the image.

The rest of the paper classified as: the Section II gives a general framework of the proposed method, while Section III explains the generation of image hash step by step. Sections IV and V respectively show the results and comparison. Section VI gives the conclusion.

## II. General Framework of Proposed Method

We are in need of a robust method for detection of image tampering which would function even for a very small tampered area with respect to the image. Tamper detection using feature points is not at all a new concept. But generating a compact image hash is quite a difficult job. The more the number of features extracted from the image more is the hash length. As the image hash is sent along with the image, more hash length means more resource is required from the image at transmitting side. To address this problem, a completely novel approach of clustering is introduced to generate image hash. We are proposing the hash length would be defined by an experimentally detected optimal value of 'k' (number of clusters) which defines the hash length, and hence whatever be the number of features the hash length stays same. As a result, a greater number of features can be used in both transmitting and receiving side for tamper detection which would further boost the accuracy much high even for very small tampered area. The complete methodology is explained below.

At the very beginning, the features are extracted using SURF algorithm. The feature locations are mapped and then the locations are clustered using k-means clustering algorithm with an experimentally determined optimal 'k' value. The initial k numbers of centers are chosen randomly by the default 'k-means ++ algorithm'. The final cluster center locations detected by k-means clustering algorithm are sent as hash along with the image.

At the receiving end again, the features are extracted using SURF algorithm. The locations of the feature points are clustered using k-means clustering algorithm with the same optimal 'k' value. The initial k numbers of centers are chosen at the locations of the centers received as hash. The newly computed cluster centers are matched against the cluster centers received as hash to determine if the received image is authentic or not. The Fig. 1 presents a flowchart diagram of the procedure

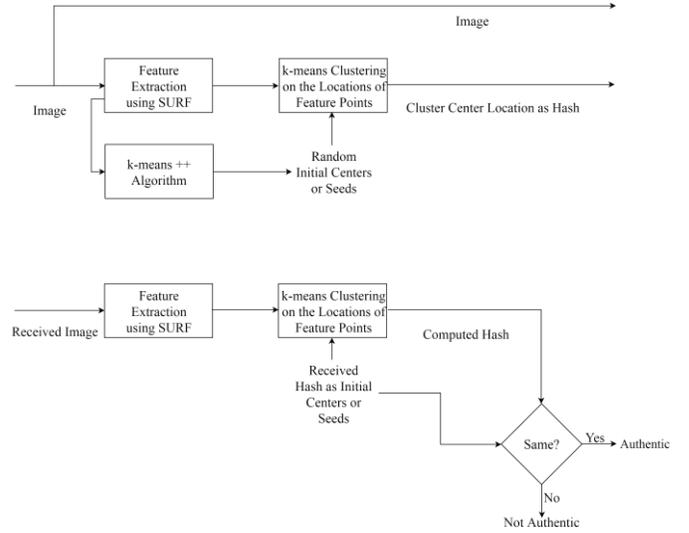

Fig. 1. Flowchart Diagram of Proposed Method

## III. Proposed Method Explained Step by Step

The method proposed for hash generation and tamper detection in this paper consists of three steps, feature extraction using SURF, k-means clustering, and comparison of the calculated hashes. In this section, we are to discuss how this is done and why it is done.

### A. Speeded Up Robust Features (SURF)

Key-point extraction is a very important operation widely used in the field of image processing for taking out the visible subject from an image. Lowe [25] suggested an algorithm, namely SIFT which seemed to outperform the previous methods providing scale and rotation invariance as well. But the problem with SIFT was its time complexity. Bay *et al.* [26] suggested a method to solve this problem, namely SURF which is quite fast to compute and keeps the repeatability and accuracy achieved by SIFT. Speeded Up Robust Features or SURF algorithm consists of several steps for detecting the feature points which are explained in the following.

- SURF uses a Hessian based blob detector to find interest points or key-points. The determinant of the Hessian matrix represents the local change around a key-point as well the extent of the response expression.

$$\mathcal{H}(x, \sigma) = \begin{bmatrix} L_{xx}(x, \sigma) & L_{xy}(x, \sigma) \\ L_{xy}(x, \sigma) & L_{yy}(x, \sigma) \end{bmatrix} \quad (1)$$

Here *L* is the convolution of second order derivative of the Gaussian with the image represented in (1). This convolution is computationally expensive to calculate and hence approximated by an integral image. An integral image is an image where each point x = (*x, y*) represents the sum of all pixels in a rectangular area formed between origin and the point x.

$$I(\mathrm{x}) = \sum_{i=0}^{i \leq x} \sum_{j=0}^{j \leq y} I(x, y) \quad (2)$$

Now any filter kernel approximation can be done using integral images by only three additions.

- The second order Gaussian kernel used for the Hessian matrix must be discretized and cropped before applying. The SURF algorithm approximates the kernels as rectangular boxes or box filters which in turn approximate the convolution quite accurately using the integral image.

$$\mathrm{Det}\ (H_{approx}) = D_{xx}D_{yy} - (wD_{xy})^2 \quad (3)$$

Here $D$ is the approximation of $L$ in the Hessian matrix. SURF uses the lowest 9×9 kernel size which represents scale σ = 1.2 and while approximating these Gaussians it is to be weighted with $w$ for the energy conservation of Gaussians in (3). Although theoretically $w$ is dependent on scale, it is more or less constant fixed at 0.9.

- To detect features across scale, several levels and octaves are examined, where SIFT scales the image down for each octave, SURF uses arbitrarily larger kernels on the same image to produce the scale space. Now each interest point detected is checked in 3×3×3 space keeping it in the center, *i.e.,* 8 neighbors surrounding it and 9 neighbors each in the level above and below. After all the 26 checks if the point is a maximum or minimum it is kept as key-point, otherwise, it is discarded.

SURF is one of the most popular interest point detectors till date as it outperforms previous methods as well as keeping the repeatability intact. For each image extracted SURF features are unique, for a visually same image the SURF features are similar while being different for a completely different image.

*B. K-means Clustering*

Data clustering is a well-known procedure used in computational intelligence and pattern recognition. Clustering means to group several objects into one or more groups or clusters in such a way that the objects in a cluster are similar to each other as well as are different from the objects present in other clusters in some respect. Data clustering is a technique which is performed by several algorithms. The clustering is done on basis of the several properties of the data points like centroid, distribution or density.

K-means clustering is a centroid based clustering algorithm where 'k' represents the number of clusters. The algorithm runs through several steps discussed in the following.

- At first 'k' centers are chosen randomly using k-means++ algorithm and assigned as initial centers. Initial centers can also be assigned by the user forcibly as well. For the hash generation at the transmitting end centers initialization is done randomly by the k-means++ algorithm while at the receiving end the seeds are given as initial cluster centers.

- Each data point is checked for each center to find out which center is closest to it and then it is assigned to that center.

- Each cluster calculates its new center using its own data points and the center is shifted. Now as the centers are shifted the distances also change and the closest center to each data point may change. Hence the first step is done again and this goes on iteratively until the centers stabilize at some specific locations.

K-means clustering is widely used in object identification problem. Here in this paper defining a value for 'k' implies to the assumption of 'k' object(s) present in the image. Using k-means on the feature point locations, *i.e.,* location of the edge or corner points, a center for each assumed object is calculated. While the image is not tampered there should be same object(s) present in the image at the receiving side also. Hence the method presented here refers to identify the same object(s) in the image for tamper detection. Due to content preserving operations, the cluster centers can change a little bit, which is due to a little difference in the feature extraction. But the centers would not change drastically. If there are some other object(s) or some objects are tampered somehow in the image the cluster centers will change massively in the image which will denote tampering. The content preserving operations and the tampering can be easily be identified by setting up a threshold value of minimum Euclidean distance as explained in the following.

*C. Comparison of Image Hash for Tamper Detection*

The image hash generated from the image at the transmitting side which is to be sent can be compared to the image hash computed from the image received to detect if the image is tampered or not. The locations of the cluster centers are hash, which completely depends on the data points generated from the feature extraction step that uses SURF algorithm. Taking the content preserving operations into account, the feature extracted from the received image may not be exactly same as the features extracted from the transmitting side image, but they would be more or less same. As the k-means clustering completely depends on the data points which are actually the locations of the feature points, any change in the number of the data points or any change in the location of extracted data points would change the computed locations of cluster centers. Hence if the features extracted from the image at receiving end are not exactly same as the features extracted from the image at transmitting end, the computed cluster center locations would not be same. This means even if the image is not tampered, the cluster centers may vary. To solve this problem, we suggest the calculation of the Euclidean distance between the two sets of cluster center locations. The minimum distance between any of the pairs is taken for the calculation. This is because even if some cluster centers are shifted a little bit the displacement would not be same for each center. The center which is shifted minimum would be very small for any content preserving operations as the extracted features would be more or less same. But in case of a tampered image, there will be feature points detected in the tampered region which would change the computed k-means cluster center locations drastically. Hence the center which is even shifted minimum would be larger compared to the shift caused by content preserving operations. So, a threshold value can easily be set up to determine if the shift of the center locations represent tampering of the image or not.

## IV. EXPERIMENTAL RESULTS

For experimentation of the method proposed in this paper a database of 200 tampered images is created by creating a patch area in the images, where for each image, the tampered area is less than 5% of the total area of the image. Another database of 100 tampered images is created from CASIA v2.0 tampered image detection evaluation database where the tampered area is more than 30%. A separate database of 4 sufficiently different images is created for several content preserving operations like JPEG compression and salt and pepper noise. In this section, we have presented plots of the accuracy in terms of number of tampered images detected out of the total number of tampered images, as well as the optimal value of 'k' and the threshold value of minimum Euclidean distance taking the content preserving operations in account.

The minimum Euclidean distance depends on the hash *i.e.*, k-means cluster center calculated for both images at transmitting end as well as receiving end. Now the k-means center completely depends on the SURF features detected within the image. Hence tampered image is identified by the features detected in the tampered region. Now the features detected in the tampered region causes the shift in the cluster centers of the k-means clustering. So more the number of features detected in the tampered region more is the shift of the cluster centers. The Fig. 2 below shows the detected SURF feature points for a set of original and its tampered pair.

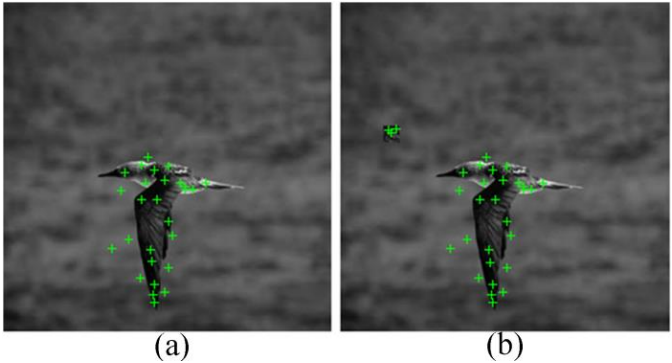

Fig. 2. SURF features detected on Images: (a) Original (b) Tampered

### A. Detection of Optimal Value of 'k'

The k-means clustering is done to minimize the hash length and to match several features at once. The value of 'k' or the number of clusters is completely experimentally determined. As we have assumed there would be 'k' object(s) in the image, practically it would not be 'k' and moreover, it would vary. Here 'k' can be chosen any arbitrary positive integer, but the experimental data shown in Fig. 3 shows there is an optimal value of 'k' to be used.

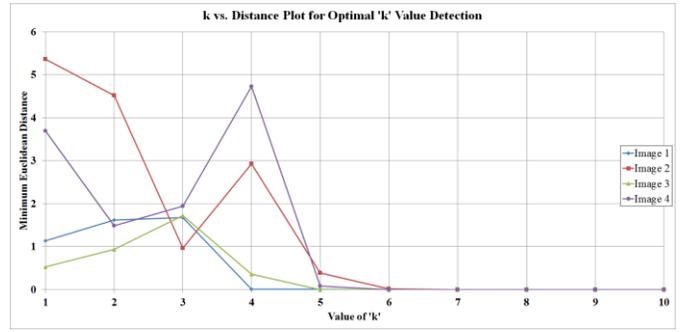

Fig. 3. K vs. Distance Plot for Detection of Optimal Value of K

It is clearly seen that for the larger values of 'k' the minimum Euclidean distance is decreasing and stabilizing at zero. At the lower values of 'k', the distances are larger. Whenever the distances are larger it is easier to distinguish the tampered images as well as easier to set a threshold up for the content preserving operations. Here from the graph, it is clearly seen that average distance of all the four images is highest at k=1. Hence, we can infer that the experimentally detected optimal value of k is 1.

### B. Determining the Threshold Distance

After k-means clustering is done, minimum Euclidean distance is used to detect the tampering. Under several content-preserving operations, the feature detected may not be exactly same, though they will be more or less same. But this small change in data points will result in a little shift of the k-means cluster centers, while for tampered images the shift will be much larger. So, it is very much required to optimally detect a threshold to distinguish the tampered images from the original ones which have gone through content preserving operations.

To determine the threshold of the minimum Euclidean distance we plot the accuracy, measured in terms of the original image detected as well as of the tampered image detected, for several possible threshold values of the minimum Euclidean distance and choose the optimal value of the threshold. Below in Fig. 4 a plot containing the plot of accuracy in terms of original image detected for salt and pepper noise and JPEG compression is shown, along with the plot of accuracy in terms of tampered image detected.

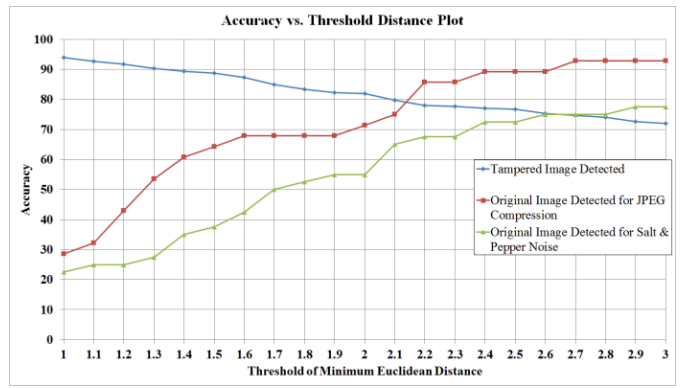

Fig. 4. Accuracy vs. Threshold Distance Graph

Here from the plot, the optimal value of the threshold for both the content preserving operations is chosen such a way when the accuracy would be same for detection of original image under content preserving operations and detection of tampered image. The optimally detected threshold of the minimum Euclidean distance is 2.1375 where both accuracy of original image detection under salt and pepper noise and accuracy of tampered image detection is 79.33%. Again, the optimally detected threshold of the minimum Euclidean distance is 2.6469 where both accuracy of original image detection under JPEG compression and accuracy of tampered image detection is 75.33%. So, we can get the average threshold by averaging the values which would be 2.3922. The accuracy for tampered image detection at the average threshold detected optimally *i.e.,* 2.3922 is 77% approximately.

*C. Tamper Detection*

The detection of tampered image is done by measuring the minimum Euclidean distance. Taking the content preserving operations into account, a threshold value is optimally calculated. If the minimum Euclidean distance measured between the hashes calculated from the image received and the received hash crosses that threshold, the image is tampered.

The image database of 200 images having the tampered area less than 5% of the total image size experimentally produces detection accuracy of 77%. A bar chart of the distance measured is given below in Fig. 5.

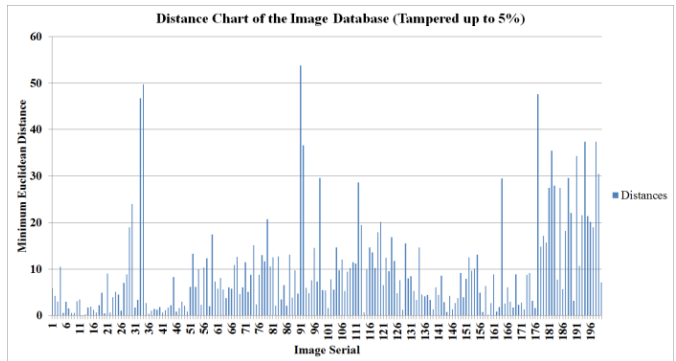

Fig. 5.   Distance Chart of the Image Database (Tampered up to 5%)

The image database of 100 images taken from CASIA v2.0 image tamper detection evaluation database, having the tampered area 30% or above of the total image size experimentally produces detection accuracy of 77%. The bar chart below in Fig. 6 shows the distance measured.

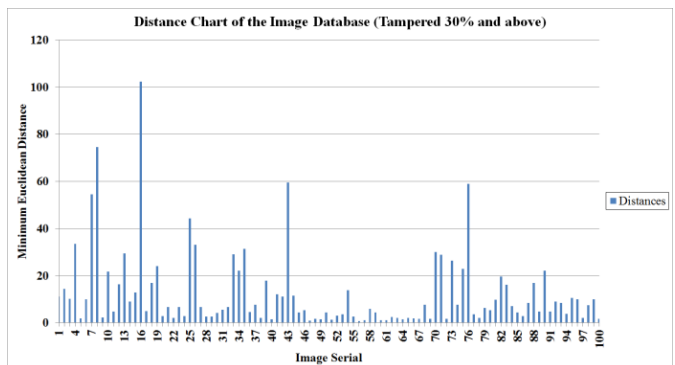

Fig. 6.   Distance Chart of the Image Database (Tampered 30% and above)

## V. COMPARISON

The method proposed in this paper outperforms the existing methods in terms of hash length. We have minimized the hash length successfully keeping the performances intact. A short comparison with the method presented in [19] is given below in Table I.

TABLE I.   COMPARISON TABLE

| Algorithm Property | Methods to be Compared | |
|---|---|---|
| | *Method in [19]* | *Proposed Method* |
| Hash Length | 634 digits | 64 bits |
| Robust against noise | Yes | Yes |
| Robust against compression | Yes | Yes |
| Detection Accuracy | 60% (Approximately) | 77% |

From Table I it is clearly visible that the method proposed in this paper has successfully brought down hash length drastically keeping the performance not even intact, but also enhancing it.

## VI. CONCLUSION

The method proposed in this paper can satisfactorily detect tamper while being robust under various content preserving operations. The hash length is minimized drastically.

The previous works on image hashing based on image features did not include all the features detected to maintain a certain hash length. As we fixed the hash length without compromising any of the image features, it is likely to achieve improved accuracy for tamper detection.

We have used the standard database for image tampering applications, and so the method is completely reliable for practical applications.

Moreover, tamper detection where the tampered area is as low as 5% of the image size is never attempted before. We used our own created image database as no standard database is available for tamper detection at this low tampered area.

ACKNOWLEDGMENT

The authors would like to acknowledge research scholars of the Department of Electronics & Communication Engineering of National Institute of Technology Silchar, India for providing support and necessary facilities for carrying out this work. The authors are thankful to Mr. Susmit Nanda, TATA Engineering & Industrial Services, for his constructive criticism and suggestions.

REFERENCES

[1] Khelifi, Fouad, and Jianmin Jiang. "Perceptual image hashing based on virtual watermark detection," *IEEE Transactions on Image Processing*, Apr. 2010, vol. 19, no. 4, pp. 981-994.

[2] R. Venkatesan, S. M. Koon, M. H. Jakubowski, and P. Moulin, "Robust image hashing," in *Proc. Int. Conf. Image Process.*, Sep. 2000, vol. 3, pp. 664-666

[3] Qiang Ma, *et al.* "Robust image authentication via locality sensitive hashing with core alignment," *Multimedia Tools and Applications*, 2017, pp. 1-22.

[4] S. S. Kozat, R. Venkatesan, and M. K. Mihcak, "Robust perceptual image hashing via matrix invariants," in *Proc. Int. Conf. Image Process. (ICIP)*, Oct. 2004, vol. 5, pp. 3443–3446.

[5] V. Monga and M. K. Mihçak, "Robust and secure image hashing via non-negative matrix factorizations," *IEEE Trans. Inf. Forensics Security*, Sep. 2007, vol. 3, no. 2, pp. 376-390

[6] Z. Tang, S. Wang, X. Zhang, W. Wei, and S. Su, "Robust image hashing for tamper detection using non-negative matrix factorization," *J. Ubiquitous Converg. Technol.*, May 2008, vol. 2, no. 1, pp. 18–26.

[7] S. Roy and Q. Sun, "Robust hash for detecting and localizing image tampering," in *Proc. IEEE Int. Conf. Image Process. (ICIP)*, Sep./Oct. 2007, pp. VI-117–VI-120.

[8] F. Ahmed, M. Y. Siyal, and V. U. Abbas, "A secure and robust hash-based scheme for image authentication," *Signal Process.*, May 2010, vol. 90, no. 5, pp. 1456–1470.

[9] Cai-Ping Yan, Chi-Man Pun, and Xiao-Chen Yuan, "Quaternion-based image hashing for adaptive tampering localization," *IEEE Transactions on Information Forensics and Security*, Dec. 2016, vol. 11, no. 12, pp. 2664-2677.

[10] V. Monga, D. Vats, and B. L. Evans, "Image authentication under geometric attacks via structure matching," in *Proc. IEEE Int. Conf. Multimedia Expo (ICME)*, Jul. 2005, pp. 229–232.

[11] Junlin Ouyang, Yizhi Liu, and Huazhong Shu, "Robust hashing for image authentication using SIFT feature and quaternion Zernike moments," *Multimedia Tools and Applications*, Jan. 2017, vol. 76, no. 2, pp. 2609-2626.

[12] Yan Zhao, Shuozhong Wang, Xinpeng Zhang, and Heng Yao, "Robust hashing for image authentication using Zernike moments and local features" *IEEE transactions on information forensics and security*, Jan. 2013, vol. 8, no. 1, pp. 55-63.

[13] Saba A. Shaikh, and Samadhan A. Sonavane, "A Novel Approach For Image Hashing Using Ring Partition and Invariant Vector Distance," *International Journal on Emerging Trends in* Technology, Jul. 2017, vol. 1, no. 2.

[14] S. Battiato, G. M. Farinella, E. Messina, and G. Puglisi, "Robust image alignment for tampering detection," *IEEE Trans. Inf. Forensics Security*, Aug. 2012, vol. 7, no. 4, pp. 1105–1117.

[15] W. Lu and M. Wu, "Multimedia forensic hash based on visual words," in *Proc. 17th IEEE Int. Conf. Image Process. (ICIP)*, Sep. 2010, pp. 989-992

[16] Xudong Lv, and Z. Jane Wang. "Perceptual image hashing based on shape contexts and local feature points," IEEE Transactions on Information Forensics and Security, Jun. 2012, vol. 7, no. 3, pp. 1081-1093.

[17] Zhenjun Tang, Liyan Huang, Xianquan Zhang, and Huan Lao, "Robust image hashing based on color vector angle and canny operator," *AEU-International Journal of Electronics and Communications*, Jun. 2016, vol. 70, no. 6, pp. 833-841.

[18] X. Wang, K. Pang, X. Zhou, Y. Zhou, L. Li, and J. Xue, "A visual model-based perceptual image hash for content authentication," *IEEE Trans. Inf. Forensics Security*, Jul. 2015, vol. 10, no. 7, pp. 1336–1349.

[19] Chi-Man Pun, Cai-Ping Yan, and Xiao-Chen Yuan, "Image Alignment-Based Multi-Region Matching for Object-Level Tampering Detection," *IEEE Transactions on Information Forensics and Security*, Feb. 2017, vol. 12, no. 2, pp. 377-391.

[20] Cai-Ping Yan, Chi-Man Pun, and Xiao-Chen Yuan, "Multi-scale image hashing using adaptive local feature extraction for robust tampering detection," *Signal Processing*, Apr. 2016, vol. 121, pp. 1-16.

[21] Vishal Monga, and Brian L. Evans, "Perceptual image hashing via feature points: performance evaluation and tradeoffs," *IEEE Transactions on Image Processing*, Nov. 2006, vol. 15, no. 11, pp. 3452-3465.

[22] Ashwin Swaminathan, Yinian Mao, and Min Wu, "Robust and secure image hashing," *IEEE Transactions on Information Forensics and security*, Jun. 2006, vol. 1, no. 2, pp. 215-230.

[23] Chi-Man Pun, Cai-Ping Yan, and Xiao-Chen Yuan, "Robust image hashing using progressive feature selection for tampering detection," *Multimedia Tools and Applications*, 2017, pp. 1-25.

[24] Vishal Monga, and Brian L. Evans, "Robust perceptual image hashing using feature points," *Image Processing, International Conference on*, Oct. 2004, vol. 1, pp. 677-680.

[25] David G. Lowe, "Distinctive image features from scale-invariant keypoints," *International journal of computer vision*, Nov. 2004, vol. 60, no. 2, pp. 91-110.

[26] H. Bay, A. Ess, T. Tuytelaars, and L. Van Gool, "SURF:Speeded Up Robust Features," *Computer Vision and Image Understanding (CVIU)*, Jun. 2008, vol. 110, no. 3, pp. 346–359.

[27] David Arthur, and Sergi Vassilvitskii, "K-means++: The Advantages of Careful Seeding," *Proc. of the Eighteenth Annual ACM-SIAM Symposium on Discrete Algorithms*, Jan. 2007, pp. 1027–1035.

[28] Stuart P. Lloyd, "Least Squares Quantization in PCM," *IEEE Transactions on Information Theory*, Mar. 1982, vol. 28, no. 2, pp. 129–137.

[29] G. A. F. Seber, *Multivariate Observations*. Hoboken, NJ: John Wiley & Sons, Inc., Sep. 2009, vol. 252.

[30] H. Spath, *Cluster Dissection and Analysis: Theory, FORTRAN Programs, Examples*. Transl. J. Goldschmidt. New York: Halsted Press, 1985 [Ellis Horwood Ltd Wiley, Chichester 1985, vol. 226 pp. 182-182].

[31] *CASIA Image Tampering Detection Evaluation Database V2.0* [Online] Available: http://forensics.idealtest.org/